\documentclass[preprint]{article} 
\usepackage{aaai18}  
\usepackage{times}  
\usepackage{helvet}  
\usepackage{courier}  
\usepackage{url}  
\usepackage{graphicx}  
\copyrighttext {This is the authors' draft. The final version was published in AAAI 2018 (Student Abstract and Poster Program).}
 
\begin{document}
%
\title{Learning Feature Representations for Keyphrase Extraction}
\author{
	Corina Florescu \and Wei Jin\\
	{CorinaFlorescu@my.unt.edu, wei.jin@unt.edu}\\
}
\maketitle

\begin{abstract}
	
In supervised approaches for keyphrase extraction, a candidate phrase is encoded with a set of hand-crafted features and machine learning algorithms are trained to discriminate keyphrases from non-keyphrases. Although the manually-designed features have shown to work well in practice, feature engineering is a difficult process that requires expert knowledge and normally does not generalize well. In this paper, we present SurfKE, a feature learning framework that exploits the text itself to automatically discover patterns that keyphrases exhibit. Our model represents the document as a graph and automatically learns feature representation of phrases. The proposed model obtains remarkable improvements in performance over strong baselines.

\end{abstract}

\section{Introduction}

Keyphrase extraction (KE) is the task of automatically extracting descriptive phrases or concepts that represent the main topics of a document. Keyphrases associated with a document have shown to improve many natural language processing and information retrieval tasks. Due to their importance, numerous approaches to KE have been proposed along two lines of research: supervised and unsupervised.

In supervised approaches, a candidate phrase is encoded with a set of features, such as its  {\em tf-idf}, position in the document or part-of-speech tag, and machine learning algorithms are trained to classify it as either positive (i.e., keyphrases) or negative (i.e., non-keyphrases). The supervised line of research has mainly been concerned with the design of various hand-crafted features which may yield better classifiers that more accurately predict the keyphrases of a document. 
For example, \citeauthor{hulth2003improved} \shortcite{hulth2003improved} used a combination of lexical and syntactic features, such as the collection frequency and the part-of-speech tag of a phrase; \citeauthor{medelyan2009human} \shortcite{medelyan2009human} designed features to include the spread of a phrase and Wikipedia-based keyphraseness.
Although these features have shown to work well in practice, many of them are computed based on statistical information collected from the training documents which are less suited outside of that domain.  Thus, it is essential to automatically discover such features or representations without relying on feature engineering which normally does not generalize well. 

An intuitive approach to replace the process of feature design is to use the existing neural language models \cite{mikolov2013linguistic}.
Learning word representations from large collections of text is an efficient way to capture rich semantic and syntactic information about words. However, even if these representations may carry a lot of information, they are not effective enough if directly used for KE.
Similar to the design of hand-crafted features which are built on various observations, a feature learning process should exploit the structure of the text itself to discover patterns that keyphrases exhibit. 
Recently, \citeauthor{perozzi2014deepwalk} \shortcite{perozzi2014deepwalk} introduced a generalization of the neural language models that allows for the exploration of a network structure through a stream of short random walks. Applying a similar line of reasoning to word graphs built from natural language documents results in representations that capture characteristics of words in relation to the topology of the document.

Inspired by this work, we propose SurfKE, a supervised approach to KE that represents a document as a word graph and learns feature representations of graph nodes. Our experiments on a synthesized dataset show that: (1) our approach has the potential to automatically learn informative features for KE; (2) our model that uses its self-discovered features obtains better results than strong baselines.

\section{Proposed Model}
\label{sec:methods}

Our model involves three essential steps, as detailed below.

\noindent		
{\bf{Graph Construction.}} \label{graph_construction}
Let $d_t$ be a target document for extracting keyphrases. 
We build an undirected word graph $G = (V, E)$, where each unique word that passes the part-of-speech filters corresponds to a vertex $v_i \in V$.
Two vertices $v_i$ and $v_j$ are linked by an edge $(v_i,v_j) \in E $ if the words corresponding to these vertices co-occur within a window of $w$ contiguous tokens in $d_t$.
The weight of an edge (denoted as $w_{ij}$)  
is computed based on the co-occurrence count of the two words within a window of $w$ successive tokens in $d_t$. 

\noindent	
{\bf{Learning Feature Representations.}}	
Let $G$ be an undirected graph constructed as above. Our goal is to learn a 
mapping function $f: V \rightarrow {\rm I\!R}^{|V| \times d}$, $d \leq$ $|V|$ where $d$ is a parameter specifying the number of dimensions for the embeddings. This function $f$ represents the latent representations associated with each vertex $v_i \in V$. 
Similar to neural language models, the network feature learning  model needs a corpus and a vocabulary in order to learn a mapping of nodes to a low-dimensional space of features.
We define our vocabulary as the set of graph vertices, where the graph is built from the target document. 
To generate a corpus for our algorithm, we leverage a biased sampled random walk strategy.
More precisely, let us consider an arbitrary node $v_i \in V$ and $\Omega_{v_i}$ = $v_{i_1}, v_{i_2}, ..., v_{i_l}$ a biased random walk of length $l$ starting at vertex $v_i$ ($v_{i_1} =v_i$).
Each node $v_{i_j} \in \Omega_{v_i}$, $j=2,l$ is generated by the following transition probability distribution:

$ \pi = P(v_{i_j}=x|v_{i_{j-1}}=y) = \left\{
\begin{array}{ll}
\frac{w_{xy}}{T} & (x, y) \in E \\
0 & \mbox{otherwise}
\end{array}
\right.
$

\noindent 
where $w_{xy}$ is the weight of edge $(x,y) \in E$ and $T$ is the sum of weights over all edges in the graph.
Hence, for each node $v_i \in V$, we sample a fixed number of biased random walks with respect to the weight of edges and use them as contexts to learn the distributed representation of words.	

\noindent	
{\bf{Candidate Phrases.}}		
Candidate words that have contiguous positions in a document are concatenated into phrases. The feature vector for a multi-word phrase is obtained by taking the mean of the vectors of words constituting 
the phrase.

\section{Experiments and Results}

{\bf{Datasets and Evaluation Measures.}}	
To evaluate the performance of our model, we carried out experiments on a synthesized dataset including news articles and research papers from several domains. 
The synthesized dataset contains 1308 documents collected from two benchmark datasets, DUC \cite{wan2008single} and Inspec \cite{hulth2003improved}, and 
a set of documents from the MEDLINE/PubMed database. 
The human assigned keyphrases available for each dataset were used as gold standard for evaluation.

We measure the performance of our model by computing Precision, Recall and F1-score. The evaluation metrics were averaged in a 10-fold cross-validation setting where the training and test sets were created at the document level.   
The model parameters such as the vector dimension or the number of walks per node were estimated on a validation set. 
We used the Gaussian Na\"ive Bayes classifier implemented in the scikit-learn library to train our model. Similar to our baselines, we evaluate the top 10 predicted keyphrases returned by the model, where candidates are ranked based on the confidence scores as output by the classifier.

{\bf Results and Discussion.}
We compare SurfKE with two supervised models, KEA \cite{frank1999domain} and Maui \cite{medelyan2009human}, 
and two unsupervised models, KPMiner \cite{el2010kp} and PositionRank \cite{florescu2017positionrank}.
KEA is a representative system for keyphrase extraction with the most important features being the frequency and the position of a phrase in a document.  
Maui is an extension of KEA which includes features such as keyphraseness, the spread of a phrase and statistics gathered from Wikipedia.
KPMiner uses statistical observations to limit the number of candidates and ranks phrases using the {\em tf-idf} model in conjunction with a boosting factor which aims at reducing the bias towards single word terms. 
PositionRank is an unsupervised graph-based model, that incorporates the position information of a word's occurrences into a biased PageRank to score words that are later used to score and rank candidate phrases.

Table \ref{results} shows the comparison of SurfKE with the four baselines described above. As we can see in the table, SurfKE substantially outperforms all baseline approaches on the synthesized dataset. 
For instance, SurfKE obtains an F1-score of $0.260$ compared to $0.240$ and $0.235$ achieved by the best performing baselines, PositionRank and Maui, respectively. 
With relative improvements of 50.68\% 33.33\%, 15.18\% and 8.91\% over KEA, KPMiner, Maui and PositionRank respectively, the Precision of our model is
significantly superior to that of the other models.  

\begin{table}
	\centering
	\begin{tabular}{|l|c|c|c|c|}
		\hline
		Approach &  Precision &  Recall &  F1-score \\
		\hline
		KEA &  0.146&  0.296&  0.179 \\
		Maui &  0.191 &  0.375  &  0.235 \\
		KpMiner &  0.165 &  0.357  &  0.220 \\
		PositionRank & 0.202 & 0.362 & 0.240 \\
		SurfKE &  \bf 0.220&  \bf 0.390 &  \bf 0.260 \\			
		\hline
	\end{tabular}	
	\caption{The comparison of SurfKE with baselines.}  \label {results}	
\end{table}

\section{Conclusion and Future Work}

We propose SurfKE, a supervised model for KE that represents the target document as a word graph and uses a biased sampled random walk model to generate short sequences of nodes which are later used as contexts to learn the node representations. 
Our experiments show that (1) our approach has the potential to exploit the word graph to capture those "central" terms that represent the text; (2) SurfKE obtains remarkable improvements in performance over strong baselines for KE. 		
In future work, we would like to explore other potential neighborhoods of a node for guiding random walks, e.g., those words that appear early in the document or more topically relevant to the text. 



\fontsize{9.0pt}{10.0pt} \selectfont
\bibliography{aaai18_bibfile}

\begin{thebibliography}{}

\bibitem[\protect\citeauthoryear{El-Beltagy and Rafea}{2010}]{el2010kp}
El-Beltagy, S.~R., and Rafea, A.
\newblock 2010.
\newblock Kp-miner: Participation in semeval-2.
\newblock In {\em SemEval'10}.

\bibitem[\protect\citeauthoryear{Florescu and
  Caragea}{2017}]{florescu2017positionrank}
Florescu, C., and Caragea, C.
\newblock 2017.
\newblock Positionrank: An unsupervised approach to keyphrase extraction from
  scholarly documents.
\newblock In {\em ACL'17},  1105--1115.

\bibitem[\protect\citeauthoryear{Frank \bgroup et al\mbox.\egroup
  }{1999}]{frank1999domain}
Frank, E.; Paynter, G.~W.; Witten, I.~H.; Gutwin, C.; and Nevill-Manning, C.~G.
\newblock 1999.
\newblock Domain-specific keyphrase extraction.
\newblock In {\em IJCAI'99},  668--673.

\bibitem[\protect\citeauthoryear{Hulth}{2003}]{hulth2003improved}
Hulth, A.
\newblock 2003.
\newblock Improved automatic keyword extraction given more linguistic
  knowledge.
\newblock In {\em EMNLP'03}.

\bibitem[\protect\citeauthoryear{Medelyan, Frank, and
  Witten}{2009}]{medelyan2009human}
Medelyan, O.; Frank, E.; and Witten, I.~H.
\newblock 2009.
\newblock Human-competitive tagging using automatic keyphrase extraction.
\newblock In {\em EMNLP'09},  1318--1327.

\bibitem[\protect\citeauthoryear{Mikolov, Yih, and
  Zweig}{2013}]{mikolov2013linguistic}
Mikolov, T.; Yih, W.-t.; and Zweig, G.
\newblock 2013.
\newblock Linguistic regularities in continuous space word representations.
\newblock In {\em HLT-NAACL}, volume~13,  746--751.

\bibitem[\protect\citeauthoryear{Perozzi, Al-Rfou, and
  Skiena}{2014}]{perozzi2014deepwalk}
Perozzi, B.; Al-Rfou, R.; and Skiena, S.
\newblock 2014.
\newblock Deepwalk: Online learning of social representations.
\newblock In {\em KDD},  701--710.

\bibitem[\protect\citeauthoryear{Wan and Xiao}{2008}]{wan2008single}
Wan, X., and Xiao, J.
\newblock 2008.
\newblock Single document keyphrase extraction using neighborhood knowledge.
\newblock In {\em AAAI'08},  855--860.

\end{thebibliography}
\bibliographystyle{aaai}

\end{document}